\documentclass[10pt,twocolumn,letterpaper]{article}

\usepackage[pagenumbers]{cvpr} 

\usepackage{graphicx}
\usepackage{amsmath}
\usepackage{amssymb}
\usepackage{booktabs}

\usepackage{blindtext}
\usepackage{multirow}
\usepackage[table,xcdraw]{xcolor}
 \usepackage{caption}
 \usepackage{overpic}
 \usepackage{pifont}
\newcommand{\cmark}{\ding{51}}%
\newcommand{\xmark}{\ding{55}}%
\usepackage{soul}

\usepackage[pagebackref,breaklinks,colorlinks]{hyperref}

\usepackage[capitalize]{cleveref}
\crefname{section}{Sec.}{Secs.}
\Crefname{section}{Section}{Sections}
\Crefname{table}{Table}{Tables}
\crefname{table}{Tab.}{Tabs.}

\def\cvprPaperID{80} 
\def\confName{CVPR}
\def\confYear{2023}

\begin{document}

\title{Bicubic++: Slim, Slimmer, Slimmest \\ Designing an Industry-Grade Super-Resolution Network}

\author{Bahri Batuhan Bilecen \qquad Mustafa Ayazoglu \\
Aselsan Research\\
Ankara, Türkiye\\
{\tt\small \{batuhanb,mayazoglu\}@aselsan.com.tr}
}

\newcommand{\insertFigTitle}[2]{\begin{overpic}[trim=0 0 0 0, clip,  scale=0.25]{#1}
    \put(2,2){\color{white}\footnotesize\textit{#2}}
\end{overpic}}


        

\twocolumn[{%
\renewcommand\twocolumn[1][]{#1}%
\maketitle

\begin{center}
    \centering
    \footnotesize
    \captionsetup{type=figure}
    \addtolength{\tabcolsep}{-5pt} 
     \begin{tabular}{cccc}
     Portion of LR & Bicubic (1 ms) & \textbf{Ours (2.9 ms)} & ESPCN (9.6 ms)~\cite{D2S} \\ 
     \includegraphics[height=0.09\linewidth]{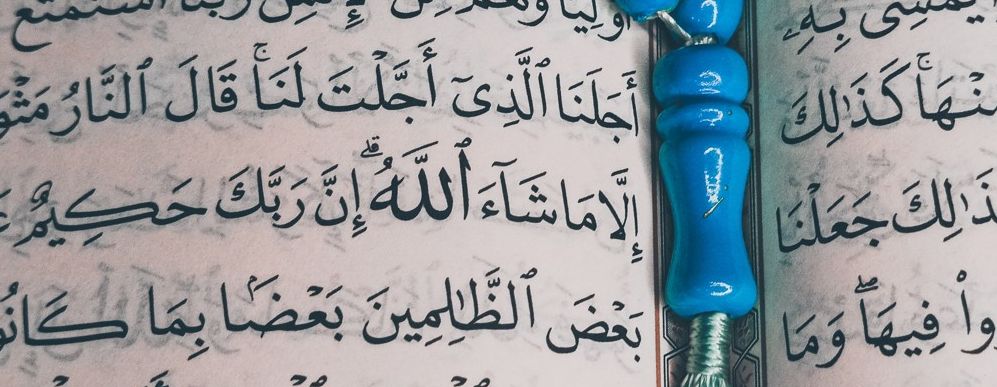}& 
        \includegraphics[height=0.09\linewidth]{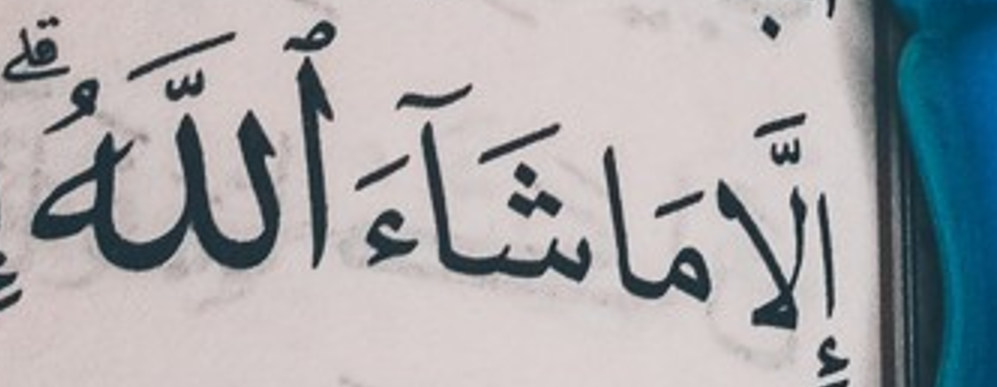}& 
        \includegraphics[height=0.09\linewidth]{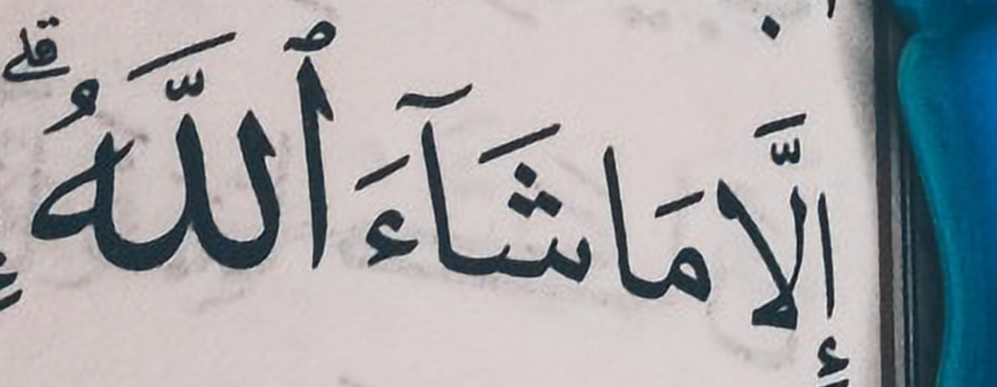}&
        \includegraphics[height=0.09\linewidth]{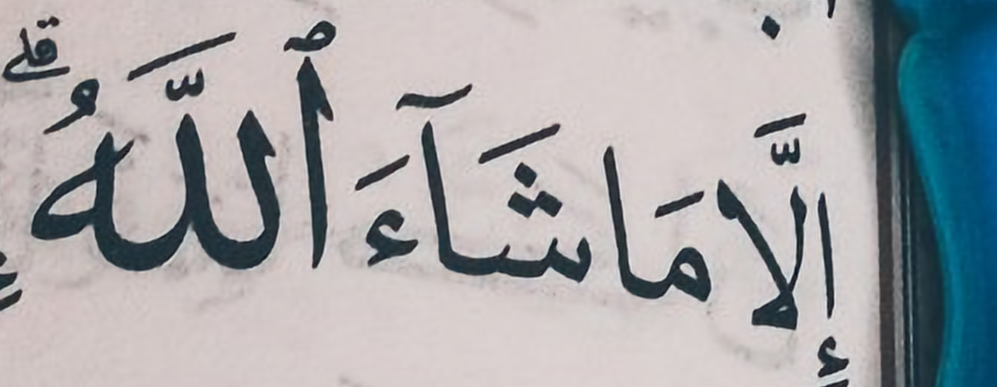}\\ 
        \includegraphics[height=0.09\linewidth]{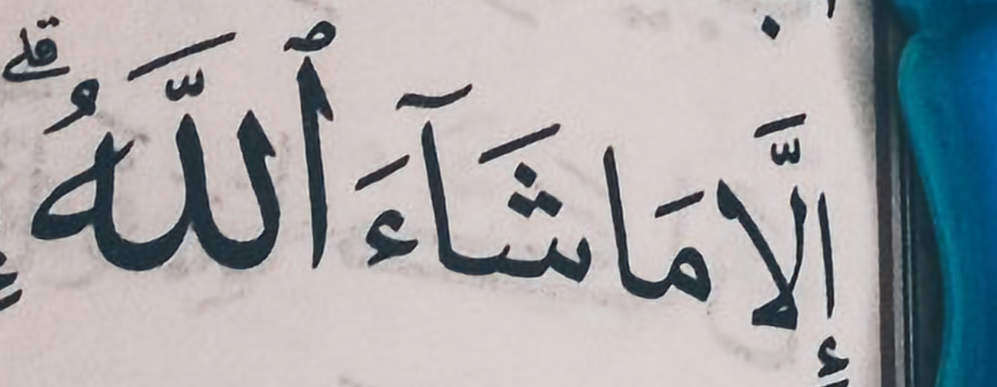}& 
        \includegraphics[height=0.09\linewidth]{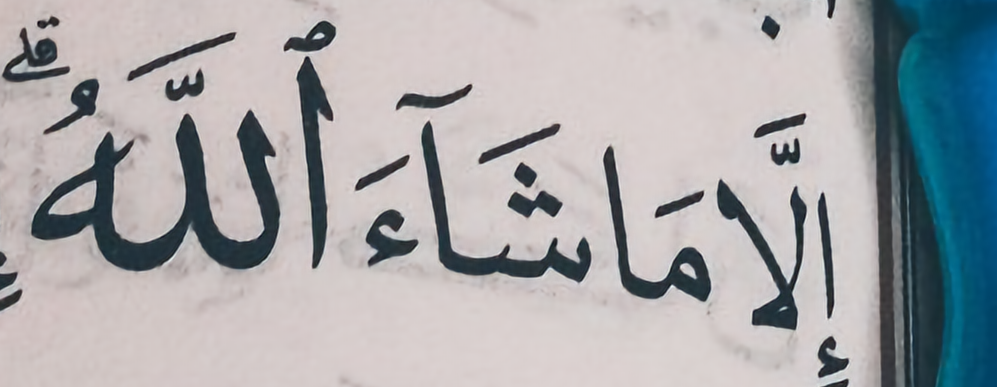}& 
        \includegraphics[height=0.09\linewidth]{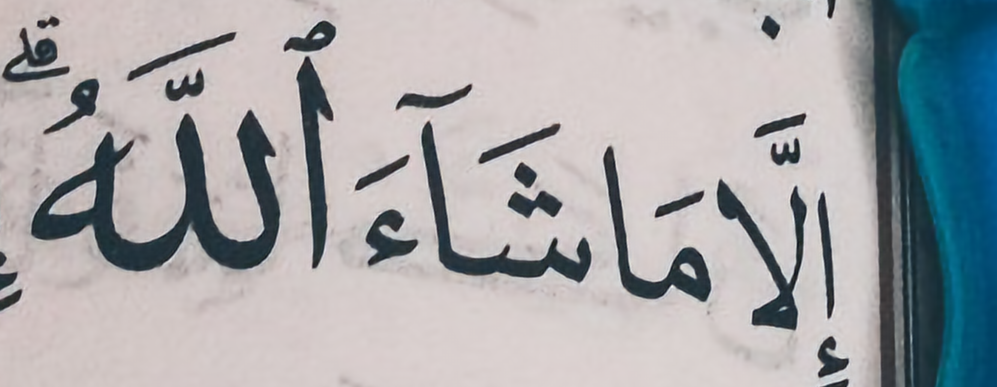}&
        \includegraphics[height=0.09\linewidth]{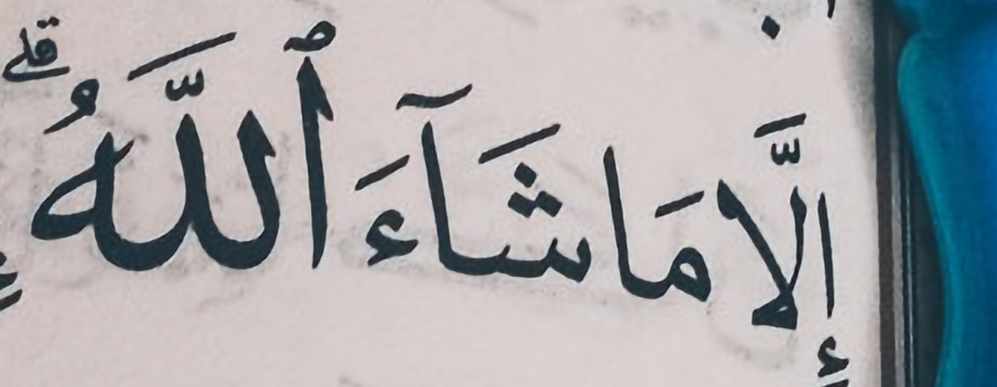}
        \\
      XCAT (11.7 ms)~\cite{xcat} & ABPN (15.8 ms)~\cite{abpn} & XLSR (25.3 ms)~\cite{xlsr} & FSRCNN (33.8 ms)~\cite{fsrcnn}
        
     \end{tabular}
    \captionof{figure}{Image \textit{70} from the NTIRE 2023 RTSR Challenge~\cite{conde2023ntire_rtsr} Track 2 $\times$3 test set~\cite{zamfir2023rtsr}, where the data is JPEG Q=90 degraded. Our practical method runs much faster ($<$ 3ms) and yields nearly identical results to the other comparable ($\sim$10-30ms) models on RTX3070.}
    \label{fig:first_page}
\end{center}%
}]

\begin{abstract}
We propose a real-time and lightweight single-image super-resolution (SR) network named Bicubic++. Despite using spatial dimensions of the input image across the whole network, Bicubic++ first learns quick reversible downgraded and lower resolution features of the image in order to decrease the number of computations. We also construct a training pipeline, where we apply an end-to-end global structured pruning of convolutional layers without using metrics like magnitude and gradient norms, and focus on optimizing the pruned network's PSNR on the validation set. Furthermore, we have experimentally shown that the bias terms take considerable amount of the runtime while increasing PSNR marginally, hence we have also applied bias removal to the convolutional layers. Our method adds $\sim$1dB on Bicubic upscaling PSNR for all tested SR datasets and runs with $\sim$1.17ms on RTX3090 and $\sim$2.9ms on RTX3070, for 720p inputs and 4K outputs, both in FP16 precision. Bicubic++ won NTIRE 2023 RTSR Track 2 $\times$3 SR competition and is the fastest among all competitive methods. Being almost as fast as the standard Bicubic upsampling method, we believe that Bicubic++ can set a new industry standard.
\end{abstract}

\begin{figure}[h!]
\centering
    \includegraphics[width=0.8\linewidth]{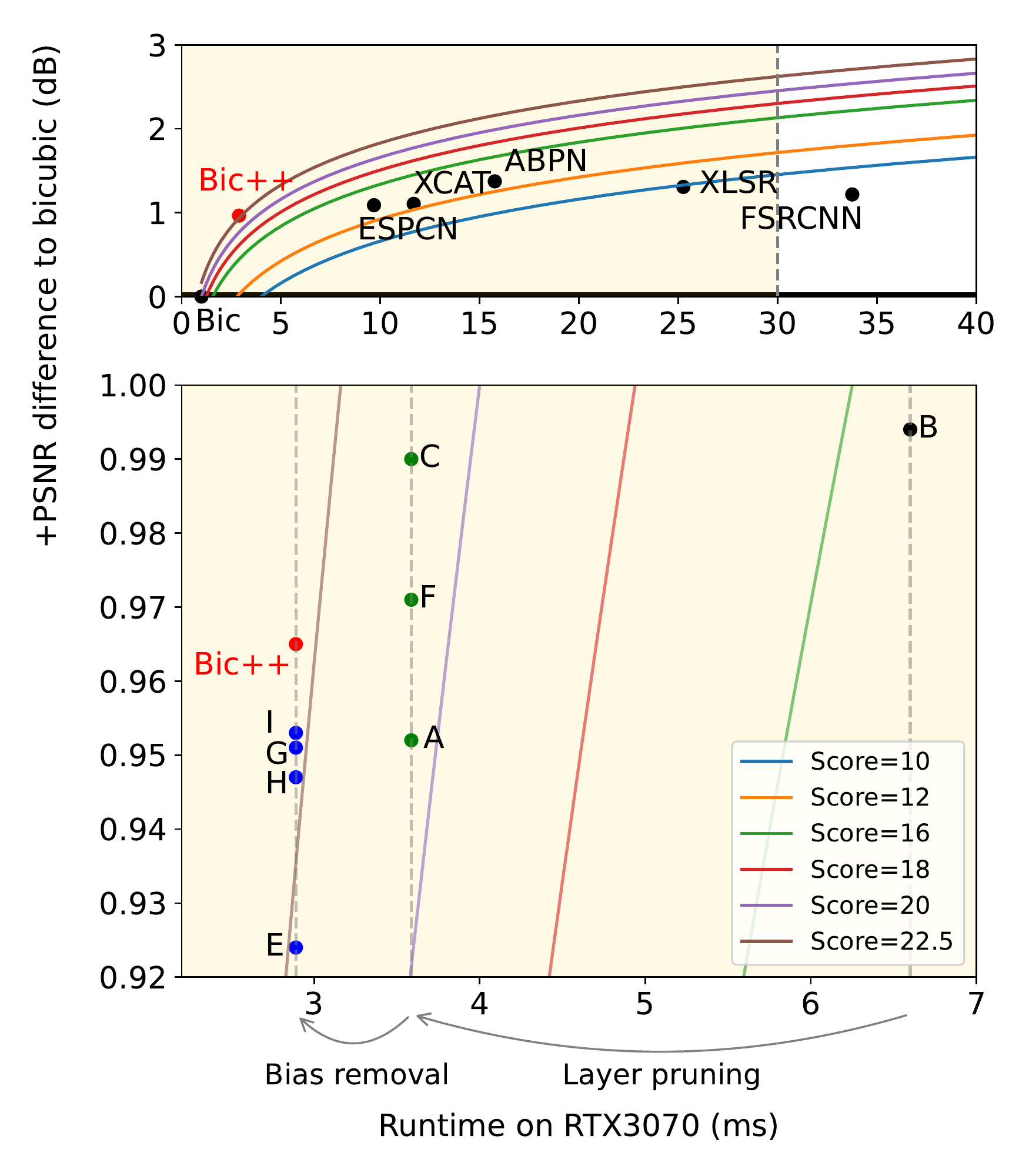}
    \caption{Runtime versus $\Delta$PSNR wrt. Bicubic, and the effect of the proposed training pipeline. Due to \cref{eqn:score}, Bicubic score is 0. Bic++ scores the highest with $<$3ms runtime, achieving $\sim$+1dB on top of Bicubic on our test set. \textbf{A}..\textbf{I} denote the models experimented. We follow  \textbf{B}$\rightarrow$\textbf{B*}$\rightarrow$\textbf{C}$\rightarrow$\textbf{Bic++}. \textbf{B*} is not in plot's range.}
    \label{fig:first_page_plot}
\end{figure}


\begin{figure*}[t]
    \centering
    \includegraphics[scale=1]{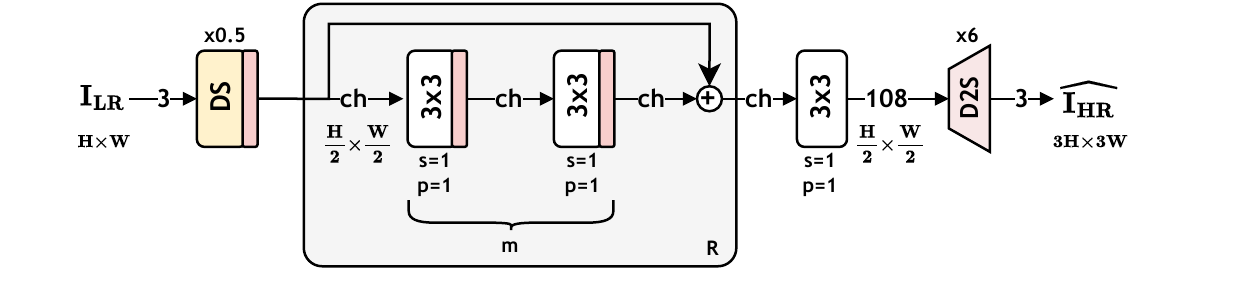}
    \caption{General network structure of Bicubic++. DS denotes the downscaling sub-network (\cref{fig:DS}), and D2S is the depth-to-space layer~\cite{D2S}. Residual addition blocks are shown by $\mathbf{R}$, and the number of convolutional layers inside $\mathbf{R}$ is $\mathbf{m}$. s and p denote the stride and padding of the convolutional layers, respectively. ch is the global channel number parameter. In the final proposed model, $\mathbf{R}$=1, $\mathbf{m}$=2, ch=32, and DS is a 3$\times$3 convolutional layer with s=2, p=1.}
    \label{fig:network}
\end{figure*}

\section{Introduction}
Single-image super-resolution (SR) is the task of upscaling a low-resolution (LR) image to a high-resolution (HR) one. SR in an highly ill-posed inverse problem, and researchers are still coming up with different approaches which would yield reasonable and high-quality HR images under varying LR conditions (noise, blur, camera motion, compression, \etc). Besides providing high-quality HR's, many SR networks also focus on runtime performance~\cite{aim2022, mai2021} since SR is also extensively utilized in real-life scenarios such as on the cameras of unmanned vehicles and surveillance systems. Nonetheless, a lot of industry-grade applications still consider Bicubic as the tried-and-true upscaling method because of its speed and ease-of-use. Despite all, we believe that a carefully constructed and trained neural network can be just as useful as the traditional Bicubic upsampling, run in real-time, and even replace Bicubic upsampling with visually more pleasing outputs.

Hence, to contribute to the ongoing development of real-time single-image super-resolution, we present our work in the \textit{NTIRE 2023 Real-Time Super-Resolution Challenge Track 2 ($\times$3 upscaling)}~\cite{conde2023ntire_rtsr,zamfir2023rtsr} and propose a lightweight Bicubic upscaling alternative, \textbf{Bicubic++}. Bicubic++ first downscales the image features by $\times$2 to reduce the number of operations significantly, and applies $\times$6 upscaling at the end. We also train the networks with our proposed three-stage training pipeline, where we first train a network with convolutional layer channels greater than the "hardware's sweet spot". Then, we apply global structured layer pruning without conditioning on weight or gradient norms while focusing on optimizing the PSNR, convolutional layer bias removal, and fine-tuning operations to further decrease its runtime speed without sacrificing much from its visual output quality. Bicubic++ ranked 1st on the \textit{NTIRE 2023 Real-Time Super-Resolution Challenge Track 2} and is the fastest method among all competitive methods.

Throughout the paper, we emphasise on optimizing the score function of the challenge given in~\cref{eqn:score}:
\begin{equation}
S(P,t)=
    \begin{cases}
        0 & \text{if } P \leq P_{bic}\\
        \frac{{}2^{P-P_{bic}}\times2}{0.1\times\sqrt{t}} & \text{else } 
    \end{cases}
    \label{eqn:score}
\end{equation}
where $P$ and $P_{bic}$ are the PSNR values of the network and the Bicubic upsampling on the test set, respectively. $t$ is the runtime of the network when a 720p LR image is given as an input and a 4K HR is obtained. In addition, due to the real-time requirements, $t$ must to be smaller than 30ms. This score formulation is provided by the competition holders and is calculated with their given code\footnote{\url{https://github.com/eduardzamfir/NTIRE23-RTSR}}.

\section{Related Works}

\textbf{Deep learning-based single-image super-resolution.}
Dong~\etal proposed SRCNN~\cite{srcnn} as the first deep learning based SR algorithm. Later on, FSRCNN~\cite{fsrcnn} was proposed, as a faster version for SRCNN. It replaced ReLU with a PReLU activation layer, reformed SRCNN with smaller filter sizes with more mapping layers, and proposed postponing the upscaling layer at the end of the network which resulted in a great speed-up. ESPCN~\cite{D2S} introduced a sub-pixel convolutional layer, often known as depth to space, which is currently used in several efficient SR networks~\cite{abpn,xcat,xlsr}. The development of deep learning-based SR was carried on by VDSR~\cite{vdsr}, EDSR~\cite{edsr}, and WDSR~\cite{wdsr} by expanding the number of parameters in exchange for accuracy and sacrificing speed. Concepts like generative adversarial networks~\cite{srgan, esrgan, gan_survey}, recursive \& residual networks~\cite{DRLN, IMDN, IMDeception}, and attention mechanism~\cite{swinIR, HAN} also rapidly took place in SR network topologies with the recent advancements~\cite{SR_survey}.

\textbf{Efficient super-resolution.} Just like the regular SR, efficient SR did not only emerge with deep learning either, as there are notable methods using sparse representations like Zeyde~\etal's~\cite{zeyde}, ANR~\cite{ANR}, and A+~\cite{A_plus}. However, thanks to the challenges and workshops held recently~\cite{mai2021, aim2022, esr_ntire_2022 ,zamfir2023rtsr, conde2023ntire_rtsr}, there is an abundance of deep learning-based SR methods now, each more efficient than the previous ones most of the time. XLSR~\cite{xlsr} utilized clipped ReLU at the end of the network to minimize the PSNR decrease in post-training INT8 quantization for the first time. ABPN~\cite{abpn} proposed an anchor-based residual network and applied quantization-aware training. SCSRN~\cite{SCSRN} applied a similar idea of residual adding to ABPN but in the feature level, and performed reparametrization of the convolutional layers to train with larger number of parameters and reduce them in the inference time. RLFN~\cite{RLFN} used three convolutional layers for residual learning to simplify feature aggregation, and stated to have achieved a balance between visual quality and runtime. A CPU-inference based method named SR-LUT~\cite{SR-LUT} proposed to train a deep SR network to construct a lookup table, and matched the input LR image patches to the output HR image patches utilizing the said table.

\section{Methodology}
\label{sec:methodology}

Our proposed network structure is given in~\cref{fig:network} in its most general form. The architecture and the design choices are inspired from the efficient SR methods~\cite{aim2022, mai2021, esr_ntire_2022}, and the experiments conducted. Based on this architecture, we tune its parameters and try different training strategies to maximize the scoring function given in~\cref{eqn:score}.

\subsection{Design Procedure and Network Architecture}
\begin{figure}[h!]
    \centering
    \includegraphics[width=1\linewidth]{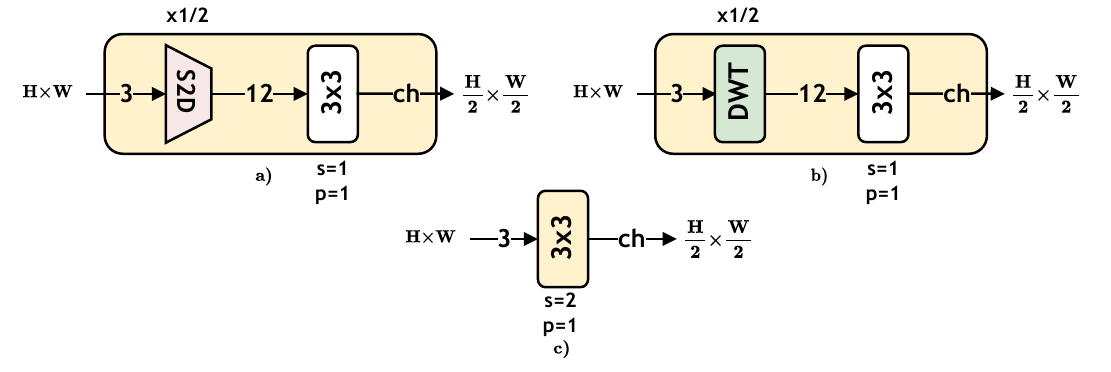}
    \caption{Different DS structures in~\cref{fig:network}. S2D (a) denotes the space-to-depth layer~\cite{D2S}, and DWT (b) is the discrete wavelet transform. We use a strided convolution (c) for DS in the end.}
    \label{fig:DS}
\end{figure}

One initial observation of ours during the design procedure was that the speed of the network does not directly depend on the number of parameters, but on the amount of activations as stated in~\cite{esr_ntire_2022}, which is a value proportional to the volume of data passing through the network. Hence, processing the feature tensors with lower width \& height than those of the input image dimensions can decrease the number of activations significantly. Furthermore, it has been shown in~\cite{adaptive_resampler} that the downscaling operation has positive effects when learned together with the upscaling. Therefore, although it seems to complicate the problem at first, we chose to "downscale" the features ($\times0.5$) by a strided convolution and then apply super-resolution operation ($\times$6) in order to speed up the network. We also experiment with other downscaling approaches (\cref{fig:DS}), but go for a strided convolution at the end. In a sense, this lower dimensional feature extraction approach can be thought as a way of compressing the data spatially; however, this compression is reversible since it is trained all together with the upscaling.

\begin{figure}[ht!]
    \centering
    \includegraphics[width=0.8\linewidth]{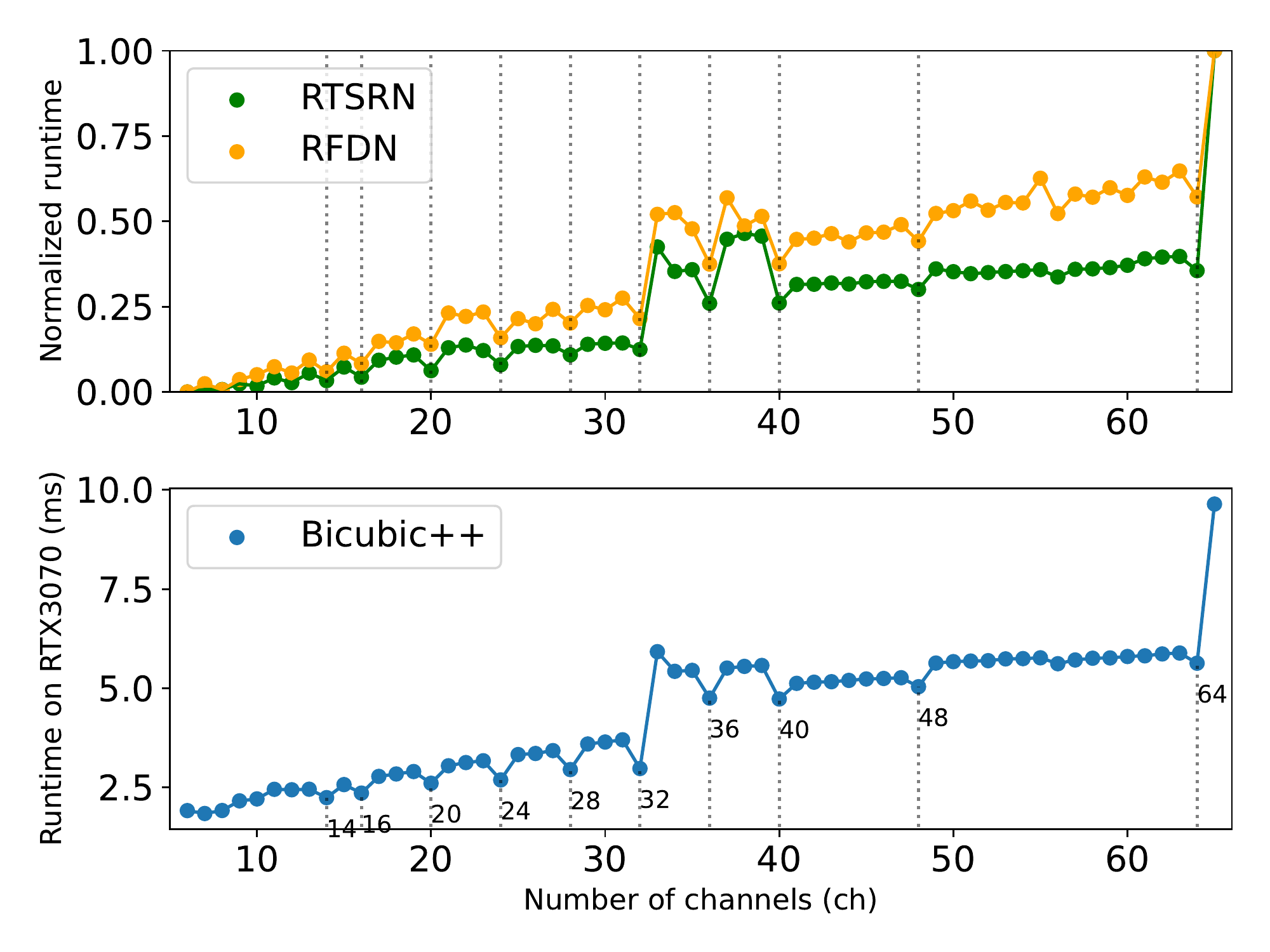}
    \caption{The effect of the channel number on the runtime and normalized runtime (runtime of ch=6 is mapped to the origin, and ch=65 to 1) of our model, RTSRN~\cite{aim2022}, and RFDN~\cite{rfdn}. Note the "sweet spots" are on ch $\in$ \{16, 20, 24, 28, 32, ...\}, and are observed in common among different models.}
    \label{fig:channel_vs_runtime}
\end{figure}

In our network, we wanted to use the "optimum" number of channels for the convolutional layers. 
We observed that decreasing the number of channels of consecutive layers does not always reflect to a decrease in the runtime. Hence, we decided to keep the same number of channels across the network (unless necessary), even though some SR methods do the quite opposite by squeeze and expand blocks~\cite{squeeze_excite} to extract features better. However, the increase in runtime caused by the blocks with varying channels outweigh its benefits considerably. To get a feeling of the optimum channel numbers, we conducted a simple experiment and observed that runtime versus the number of channels is not positively correlated all the time, and there exists some "sweet spots" (\cref{fig:channel_vs_runtime}). For our setup and under the real-time constraints, 56 channels used by RFDN~\cite{rfdn} and 28 channels used by ABPN~\cite{abpn} in the literature came out to be not the optimal points. Thus, among the options shown in~\cref{fig:channel_vs_runtime}, we opted for a channel number of 32 in our final model. However, as detailed in~\cref{subsec:ablation}, to get the most out of the model, we will start from a non-optimal channel (34) and globally prune the model to achieve 32 channels.  

In addition to all these, we also noticed that the innocent-looking bias terms of the convolutional layers are usually overlooked, and actually take a considerable portion of the overall runtime. However, the drastic decrease in the runtime caused by removing the bias terms actually overpower the marginal decrease in PSNR, which is also clearly visible from the obtained score values (\cref{fig:first_page_plot}). Hence, we wanted to eliminate the bias terms from the layers in our final network structure as well.

To sum up those aforementioned observations and ideas:
\begin{itemize}
\item Data volume processed is correlated with runtime rather than number of parameters
\item Varying the number of channels among blocks decreases runtime performance
\item There exists some "sweet spots" for the channel size for the hardware, and reaching those "sweet spots" can be done by training a larger network and pruning afterwards
\item The bias terms may take a considerable amount of time
\end{itemize}
In the end, to apply these points to our proposed network, we constructed a training pipeline. For each choice, we empirically proved its benefits. We also detail the pipeline and empirical justifications in~\cref{subsec:training_pipeline} and~\cref{subsec:ablation}, respectively.

\begin{figure*}[ht!]
    \centering
    \includegraphics[scale=1]{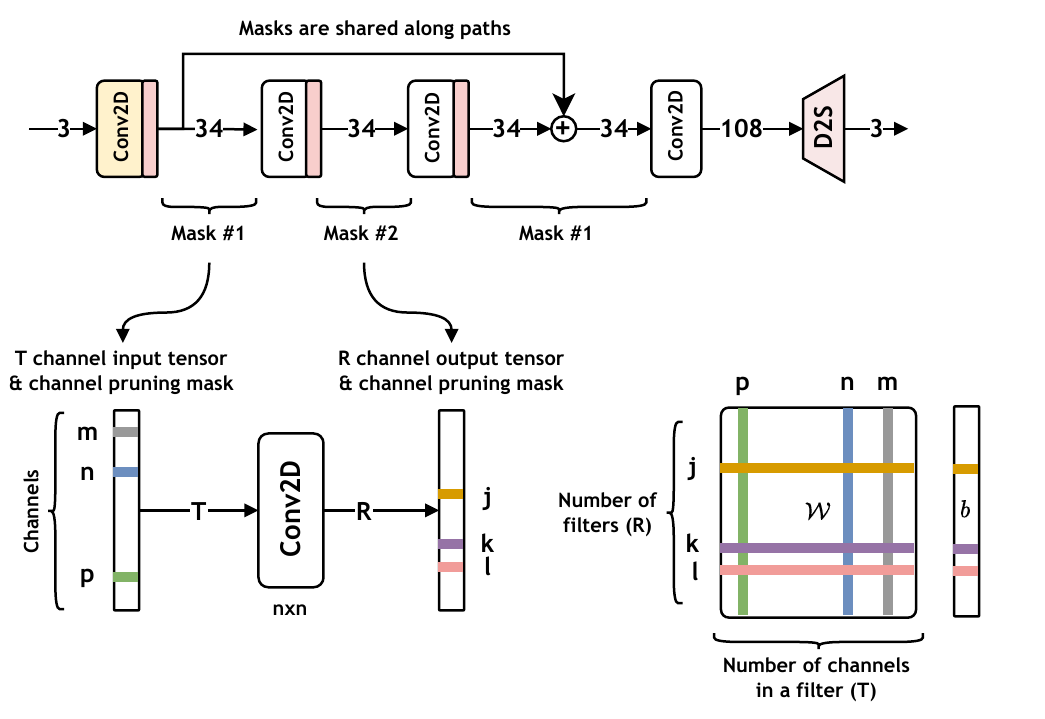}
    \caption{Global structured pruning mask generation structure. $\mathcal{W}_{R \times T}$ is the matrix consisting of the convolutional filters for a specific Conv2D layer with kernel size n$\times$n, and \textit{b} is the corresponding bias term of the layer. Each entry of $\mathcal{W}$ is an n$\times$n kernel.}
    \label{fig:pruning}
\end{figure*}

\subsection{Three-Stage Training Pipeline}
\label{subsec:training_pipeline}
We construct a three-stage training pipeline: training with a larger number of channels, global structured pruning of channels, and bias removal of convolutional layers. For the last two stages, we fine-tune the network after the modifications.

\textbf{Training hyperparameters \& computational issues:} For each stage, we use Adam optimizer with parameters $\beta_{1,2}$ 0.99 and 0.999, respectively. Models in all stages are trained for 1000 epochs, each epoch consuming 800 randomly cropped and rotated patches of dimension (108,108,3) (for LR) from JPEG Q=90 degraded DIV2K training~\cite{DIV2K_1} dataset. For the learning rate (LR), we initialize with 5e-4 and use a decaying learning rate scheduler for all stages, where for the first 500 epochs the LR stays constant and for the last 500 epochs the LR decays linearly until 1e-8.

To be able to train such a tiny model by utilizing the training hardware to its fullest, all the elements of the training pipeline need to be adjusted properly. This is due to the model forward and backward passes being done in almost no time, and memory accesses along with the validation steps becoming the actual bottleneck. To speed-up the validation, we use 48 images with same dimensions (680,452,3) (for LR) from JPEG Q=90 degraded DIV2K validation dataset, and hence be able to pick a batch size of 8 instead of 1. To take care of the memory access bottleneck, we pre-loaded both the training and the validation images to RAM and avoided fetching the data from disk continuously for each training and validation step. This way, we were able to complete a training cycle of 1000 epochs in only 1 hour using a single Tesla V100. Note that since the model's capacity is small, opting for a comparably smaller training dataset (like DIV2K) instead of favoring for a larger dataset is usually viable, and does not hurt the validation score in the end generally.


\textbf{\ul{Stage 1 - Slim: Training with a larger number of channels.}} We train the network in~\cref{fig:network}, where ch is 34, \textbf{m} is 2, \textbf{R} is 1, and DS is the strided convolutional layer in~\cref{fig:DS}. All convolutional layer biases are enabled in this stage. Respective network parameter choices are justified and explained thoroughly in~\cref{subsec:ablation}.


\textbf{\ul{Stage 2 - Slimmer: Global structured pruning.}} We take the resulting checkpoint from Stage 1 and perform the proposed pruning procedure given in~\cref{fig:pruning}. The proposed method is a structured pruning (prune channel-wise instead of kernel element-wise), in a global sense, since we prune the entire network considering the entire network's structure through sharing and transferring pruning masks and using a global fitness measure (\ie using validadion PSNR of the pruned model), instead of focusing on individual elements of the network and heuristic measures for fitness. 

In detail; for the convolutional layers to be pruned, we first construct all possible Mask \#1's in~\cref{fig:pruning} ($\mathcal{M}_1$), each only masking a single channel. After obtaining different $\mathcal{M}_1$'s, we calculate the PSNR scores on the validation set for each mask applied individually, and unify two $\mathcal{M}_1$'s (to reach the closest "sweet spot", ch=32) which cause the most marginal PSNR drop as our best \& final $\mathcal{M}_1$. 

After constructing $\mathcal{M}_1$, we move onto the Mask \#2 ($\mathcal{M}_2$). We follow the same procedure as in the creation of $\mathcal{M}_1$; however, now we do not start from the beginning and keep the $\mathcal{M}_1$'s modifications (though starting from scratch resulted the same set of masks). We do not perform any additional training while $\mathcal{M}_1$ and $\mathcal{M}_2$ are being selected, but only after $\mathcal{M}_1$ and $\mathcal{M}_2$ are all applied. We do not prune the bias terms here, either. One observation we made here was the network's overall performance was a lot more sensitive to the channels in $\mathcal{M}_1$ than those in $\mathcal{M}_2$.

\textbf{\ul{Stage 3 - Slimmest: Bias removal of the convolutional layers.}} After applying $\mathcal{M}_1$ \& $\mathcal{M}_2$ and reducing ch down from 34 to 32, we take the checkpoint from Stage 2, disable all bias from the convolutional layers, and fine-tune the overall network again.

\section{Experiments}
\label{sec:experiments}
We have performed extensive experiments trying to find the optimal parameters for our general network structure in~\cref{fig:network}, with the given scoring function in~\cref{eqn:score}. In addition to network structure  and the proposed training pipeline in~\cref{sec:methodology}, we experiment with partial loading of model components to gradually decrease the size and runtime of the network. All PSNR scores in the ablation study are evaluated on our 48-image JPEG Q=90 degraded DIV2K validation set, unless stated otherwise.

\begin{table*}[ht!]
\centering
\scriptsize
\begin{tabular}{@{}rccccccccccccc@{}}
\toprule

\multirow{2}{*}{\textbf{Exp}} & \multirow{2}{*}{\textbf{Models}} & \multirow{2}{*}{\textbf{DS}}                   & \multirow{2}{*}{\textbf{Act}}           & \multirow{2}{*}{\textbf{ch}}          & \multirow{2}{*}{\textbf{m}}           & \multirow{2}{*}{\textbf{R}}           & \multirow{2}{*}{\textbf{b}}           & \multirow{2}{*}{\begin{tabular}[c]{@{}c@{}}\textbf{PSNR (Y)} \\ \textbf{(dB)}\end{tabular}}        & \multirow{2}{*}{\begin{tabular}[c]{@{}c@{}}\textbf{Runtime} \\ \textbf{(ms)}\end{tabular}} &  \multirow{2}{*}{\textbf{Score}}      &  \multirow{2}{*}{\textbf{Load}}        &  \multirow{2}{*}{\textbf{Operation}}   & 
\\
\\

\cmidrule(lr){1-1} \cmidrule(lr){2-8} \cmidrule(lr){9-11}  \cmidrule(lr){12-14} \multirow{8}{*}{{\rotatebox[origin=c]{90}{\begin{tabular}[c]{@{}c@{}}\#1: DS, b, and \\activation\end{tabular}}}} & Q & SC & ReLU & 24 & 2 & 1 & \cmark & 30.157 & 3.19 & 19.86 & \xmark & \xmark \\  
& X & DWT & ReLU & 24 & 2 & 1 & \cmark & 30.219 & 3.59 & 19.55 & \xmark & \xmark \\ 
& Y & S2D & ReLU & 24 & 2 & 1 & \cmark & 30.172 & 3.33 & 19.64 & \xmark & \xmark\\
& Z & S2D & ReLU & 32 & 2  & 1 & \cmark & 30.269 & 3.72 & 19.88 & \xmark & \xmark\\
& W & DWT & ReLU & 32 & 2  & 1 & \cmark & 30.326 & 4.14 & 19.60 & \xmark & \xmark\\
& T & SC & ReLU & 32 & 2 & 1 & \cmark & 30.271 & 3.57 & 20.32 & \xmark & \xmark\\
& A                    & SC & LReLU & 32                   & 2                    & 1                    & \cmark                    & 30.282               & 3.58                                 & 20.43                & \xmark                    &                       \xmark       \\
& E                   & SC & LReLU                       & 32                   & 2                    & 1                    & \xmark                    & 30.254               & 2.89                                 & 22.32                & \xmark                    & \xmark          \\
\cmidrule(lr){2-14}

\multirow{5}{*}{{\rotatebox[origin=c]{90}{\begin{tabular}[c]{@{}c@{}}\#2: m \\and R\end{tabular}}}}  & E                   & SC & LReLU                       & 32                   & 2                    & 1                    & \xmark                    & 30.254               & 2.89                                 & 22.32                & \xmark                    & \xmark         \\
  & 1 & SC & LReLU   & 32   & 2   & 2   & \cmark    & 30.410   & 4.95   &  19.00 & \xmark    &  \xmark \\ 
& 2 & SC & LReLU   & 32   & 4   & 1   & \cmark   & 30.433   & 4.84  & 19.53 & \xmark    &  \xmark \\
& 3 & SC & LReLU  & 32   & 4   & 1   & \xmark   & 30.424  & 4.15 & 20.96  & \xmark    &  \xmark \\
& 4 & SC & LReLU  & 32   & 2   & 2   & \xmark   & 30.390   & 4.30   & 20.11   & \xmark    &  \xmark  \\ 

\cmidrule(lr){2-14}

\multirow{6}{*}{\rotatebox[origin=c]{90}{\begin{tabular}[c]{@{}c@{}}\#3: ch selection\\ for pruning\end{tabular}}}  & P                   & SC & LReLU                       & 35                   & 2                    & 1                    & \cmark                    & 30.293               & 6.50                                 & 15.29                & \xmark                    & \xmark          \\
& P*                   & SC & LReLU                       & 32                   & 2                    & 1                    & \cmark                    & 29.712               & 3.58                                 & 13.77                & P                   & Global structured pruning        \\
& R                   & SC & LReLU                       & 33                   & 2                    & 1                    & \cmark                    & 30.286               & 6.16                                 & 15.63                & \xmark                    & \xmark        \\
& R*                   & SC & LReLU                       & 32                   & 2                    & 1                    & \cmark                    & 30.113               & 3.58                                 & 18.44                & R                   & Global structured pruning        \\
&B \cellcolor[HTML]{EFEFEF}                   & \cellcolor[HTML]{EFEFEF}SC & \cellcolor[HTML]{EFEFEF}LReLU                        & \cellcolor[HTML]{EFEFEF}34                   & \cellcolor[HTML]{EFEFEF}2                    & \cellcolor[HTML]{EFEFEF}1                    & \cellcolor[HTML]{EFEFEF}\cmark                    & \cellcolor[HTML]{EFEFEF}30.324               & \cellcolor[HTML]{EFEFEF}6.60                                 & \cellcolor[HTML]{EFEFEF}15.50                & \cellcolor[HTML]{EFEFEF}\xmark                   &    \cellcolor[HTML]{EFEFEF}\xmark                           \\ 
& B* \cellcolor[HTML]{EFEFEF}                  & \cellcolor[HTML]{EFEFEF}SC & \cellcolor[HTML]{EFEFEF}LReLU                        & \cellcolor[HTML]{EFEFEF}32                   & \cellcolor[HTML]{EFEFEF}2                    & \cellcolor[HTML]{EFEFEF}1                    & \cellcolor[HTML]{EFEFEF}\cmark                    & \cellcolor[HTML]{EFEFEF}30.155               & \cellcolor[HTML]{EFEFEF}3.58                                 & \cellcolor[HTML]{EFEFEF}18.72               & \cellcolor[HTML]{EFEFEF}B                   &    \cellcolor[HTML]{EFEFEF}Global structured pruning                            \\ \cmidrule(lr){2-14}
 \multirow{6}{*}{{\rotatebox[origin=c]{90}{\begin{tabular}[c]{@{}c@{}}\#4: Fine tuning and\\bias removal\end{tabular}}}} &C \cellcolor[HTML]{EFEFEF}                  & \cellcolor[HTML]{EFEFEF}SC & \cellcolor[HTML]{EFEFEF}LReLU                      & \cellcolor[HTML]{EFEFEF}32                   & \cellcolor[HTML]{EFEFEF}2                    & \cellcolor[HTML]{EFEFEF}1                    & \cellcolor[HTML]{EFEFEF}\cmark                    & \cellcolor[HTML]{EFEFEF}30.320               & \cellcolor[HTML]{EFEFEF}3.58                                 & \cellcolor[HTML]{EFEFEF}20.99                & \cellcolor[HTML]{EFEFEF}B*                    & \cellcolor[HTML]{EFEFEF}Fine tuning (FT) \\
& \textcolor{red}{\textbf{(Bic++)}} \cellcolor[HTML]{EFEFEF}           & \cellcolor[HTML]{EFEFEF}SC & \cellcolor[HTML]{EFEFEF}LReLU                        & \cellcolor[HTML]{EFEFEF}32                   &\cellcolor[HTML]{EFEFEF}2                    & \cellcolor[HTML]{EFEFEF}1                    & \cellcolor[HTML]{EFEFEF}\xmark                    & \cellcolor[HTML]{EFEFEF}30.295               & \cellcolor[HTML]{EFEFEF}\textcolor{red}{\textbf{2.89}}                        & \cellcolor[HTML]{EFEFEF}\textcolor{red}{\textbf{22.96}}                & \cellcolor[HTML]{EFEFEF}C                    & \cellcolor[HTML]{EFEFEF}Bias removal + FT              \\
& F                   & SC & LReLU                        & 32                   & 2                    & 1                    & \cmark                    & 30.301               & 3.58                                 & 20.72                & A                    & Fine tuning              \\
& G                   & SC & LReLU                       & 32                   & 2                    & 1                    & \xmark                   & 30.281               & 2.89                                 & 22.74                & A                    & Bias removal + FT             \\
& H                    & SC & LReLU                       & 32                   & 2                    & 1                    & \xmark                    & 30.277               & 2.89                                 & 22.68                & E                    & Fine tuning             \\
& I                 & SC & LReLU                       & 32                   & 2                    & 1                    & \xmark                    & 30.283               & 2.89                                 & 22.77                & F                    & Bias removal + FT           \\
& J                 & SC & LReLU                       & 32                   & 2                    & 1                    & \xmark                    & 30.290               & 2.89                                 & 22.88                & H                    & Fine tuning          \\ 
\cmidrule(lr){2-14}

 \multirow{2}{*}{{\rotatebox[origin=c]{90}{\#5}}} & 5 & SC & LReLU  & 32   & 2   & 1   & \cmark  & 30.210   & 3.58   & 19.45   & 2    &  Partial load $\mathbf{m}$'s + FT \\
& 6 & SC & LReLU   & 32   & 2   & 1   & \cmark   & 30.281  & 3.58   & 20.43   &  1   &  Partial load $\mathbf{R}$'s + FT \\

\midrule
\multirow{7}{*}{{\rotatebox[origin=c]{90}{\begin{tabular}[c]{@{}c@{}}Other \\methods\end{tabular}}}} & Bicubic & - &   - &  -  &   - &  -  &  - & 29.334 & 1.0  & 0   & - & - &\\
& ESPCN~\cite{D2S} & - &   - &  -  &   - &  -  &  - & 30.419  & 9.68  & 13.67   & - &  - &\\
& XCAT~\cite{xcat} & - &   - &  -  &   - &  -  &  - &30.435  & 11.68  & 12.58   &  - &  - &\\
& ABPN~\cite{abpn} & - &   - &  -  &   - &  -  &  - & 30.703  & 15.76  & 13.05   & - & - &\\
& XLSR~\cite{xlsr} & - &   - &  -  &   - &  -  &  - & 30.637  & 25.25  & 9.84   & - & - &\\
& FSRCNN~\cite{fsrcnn} & - &   - &  -  &   - &  -  &  - & 30.547  & 33.75  & 8.00   & - & - &\\
& RFDN~\cite{rfdn} & - &   - &  -  &   - &  -  &  - & 30.921  & 159.3  & 4.77   & - & - &\\
\bottomrule

\end{tabular}
\caption{Quantitative ablation study to pick the best scoring model. $\mathbf{b}$ denotes the bias of convolutional layers, and LReLU is the leaky ReLU activation. \textbf{DS} configurations are given in~\cref{fig:DS}. We follow \textbf{B} $\rightarrow$ \textbf{B*}  $\rightarrow$  \textbf{C} $\rightarrow$ \textbf{Bic++} for the final model. PSNR scores are evaluated on the 48-image DIV2K validation dataset described in~\cref{subsec:training_pipeline}. Runtimes are measured on RTX3070.}
\label{tab:all_ablation}
\end{table*}

\subsection{Ablation Studies}
\label{subsec:ablation}
We provide an extensive ablation study in~\cref{tab:all_ablation}. 
We also divide the experiments in~\cref{tab:all_ablation} into 5 different sections for easy readibility and comparability: 

\textbf{\ul{Downscaling (DS), bias (b), and activation selection.}} From \textbf{Q-T}, we observed that using a \textit{strided convolution (SC)} is superior compared to discrete wavelet transform (DWT) and space-to-depth (S2D), even when the number of channels are smaller (notice that \textbf{Q\&Z} nearly got the same score despite \textbf{Q} having less number of channels). \textbf{T\&A} reveal that using leaky ReLU instead of ReLU can provide a significant benefit in PSNR without nearly any increase in the runtime. We can also see from \textbf{A\&E} that disabling bias terms can be advantageous to get higher scores.  

\textbf{\ul{Number of residual blocks (R) and convolutional layers (m).}} Results of \textbf{1-4} and \textbf{E} suggest that setting m=2 and R=1 yields the best score among tried model architectures. Hence, after the experiments done up to this stage and the observation in~\cref{fig:channel_vs_runtime}, we decided that for our final model, ch is 32, DS is SC, activations are leaky ReLU, and the bias terms should be disabled at some point.

\textbf{\ul{Selection of number of channels for pruning.}} This part includes the experiments done for searching the ideal number of channels for reducing to 32. We mainly focus on ch$\in$\{33,34,35\}, and apply our proposed pruning method (\cref{subsec:training_pipeline}) for \textbf{P, R, B}, and obtain \textbf{P*, R*, and B*}, respectively. The results reveal that choosing ch=34 and applying the pruning obtains the highest score, hence we move on from \textbf{B*}.



\textbf{\ul{Fine tuning and bias removal.}}
After training \textbf{B} and pruning to obtain \textbf{B*}, we fine-tune the network by training it again, and obtain \textbf{C}. The positive effect of pruning can be observed from the comparison of \textbf{C\&F}.

Afterwards, we remove the bias terms from \textbf{C}, fine-tune it once more, and obtain our proposed model, \textbf{Bicubic++}. \textbf{I}, cannot reach the final network's score, indicating the advantages of the proposed training pipeline. We also show other possible training paths ($\mathbf{A \rightarrow G}, \mathbf{E \rightarrow H}$, $\mathbf{A \rightarrow F \rightarrow I}, \mathbf{A \rightarrow E \rightarrow H \rightarrow J}$) to further point out that our training pipeline yields the best score. A visual comparison is also provided for this experimental setup in~\cref{fig:first_page_plot}.



\textbf{\ul{Extra experiments regarding partial loading of \textbf{R}\&\textbf{m}.}} We also experiment with the residual blocks (\textbf{R}), the number of convolutional layers (\textbf{m}), and partial loading models with different \textbf{R}\&\textbf{m} and provide the results. 

It is worthy to note that one can always increase the number of channels \& blocks and construct a model obtaining higher PSNR scores with slower runtime; however, we aimed to stay under $<$3ms on RTX3070, hence decided to stick with the choices made in this paper.

\subsection{Comparative Results}
\begin{figure*}[t]
    \footnotesize
    \centering
    \addtolength{\tabcolsep}{-5pt}
    \begin{tabular}{cccccccc}
     Portion of LR & Bicubic & \textbf{Ours} & ESPCN~\cite{D2S} & XCAT~\cite{xcat} & ABPN~\cite{abpn} & XLSR~\cite{xlsr} & FSRCNN~\cite{fsrcnn}  \\
        \includegraphics[height=0.105\linewidth]{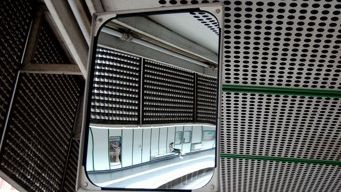}& 
        \includegraphics[width=0.105\linewidth]{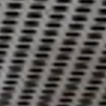}& 
        \includegraphics[width=0.105\linewidth]{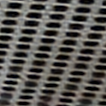}&
        \includegraphics[width=0.105\linewidth]{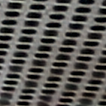}& 
        \includegraphics[width=0.105\linewidth]{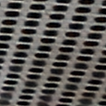}& 
        \includegraphics[width=0.105\linewidth]{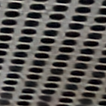}& 
        \includegraphics[width=0.105\linewidth]{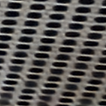}&
        \includegraphics[width=0.105\linewidth]{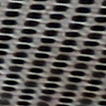}
        \\
        
        \includegraphics[height=0.105\linewidth]{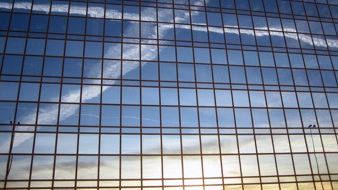}& 
        \includegraphics[width=0.105\linewidth]{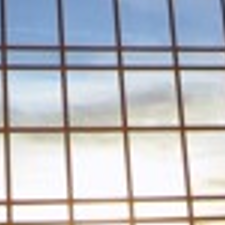}& 
        \includegraphics[width=0.105\linewidth]{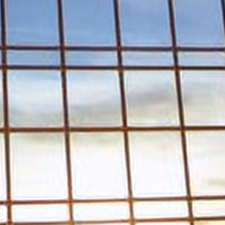}&
        \includegraphics[width=0.105\linewidth]{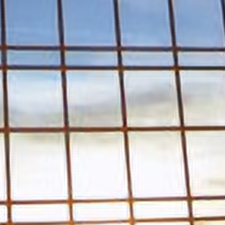}& 
        \includegraphics[width=0.105\linewidth]{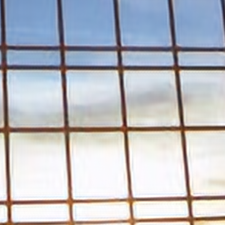}& 
        \includegraphics[width=0.105\linewidth]{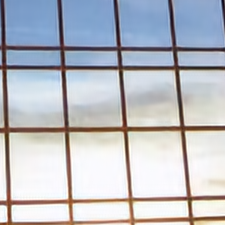}& 
        \includegraphics[width=0.105\linewidth]{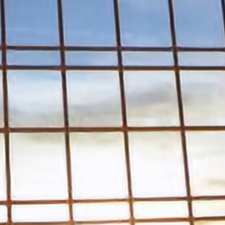}&
        \includegraphics[width=0.105\linewidth]{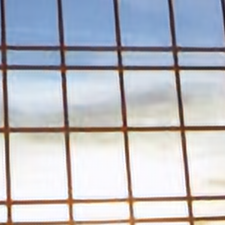}
        \\
        
        \includegraphics[height=0.105\linewidth]{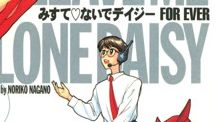}& 
        \includegraphics[width=0.105\linewidth]{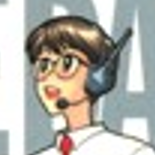}& 
        \includegraphics[width=0.105\linewidth]{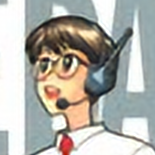}&
        \includegraphics[width=0.105\linewidth]{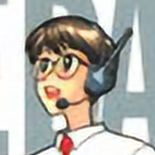}& 
        \includegraphics[width=0.105\linewidth]{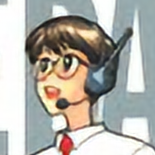}& 
        \includegraphics[width=0.105\linewidth]{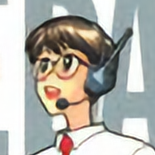}& 
        \includegraphics[width=0.105\linewidth]{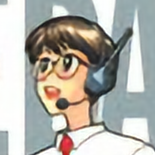}&
        \includegraphics[width=0.105\linewidth]{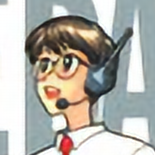}
        \\

        \includegraphics[height=0.105\linewidth]{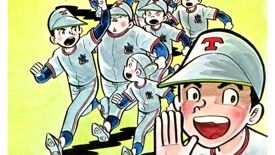}& 
        \includegraphics[width=0.105\linewidth]{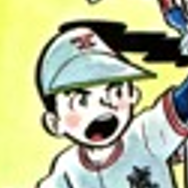}& 
        \includegraphics[width=0.105\linewidth]{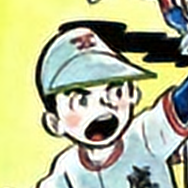}&
        \includegraphics[width=0.105\linewidth]{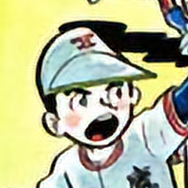}& 
        \includegraphics[width=0.105\linewidth]{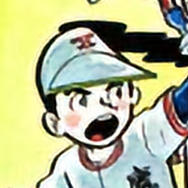}& 
        \includegraphics[width=0.105\linewidth]{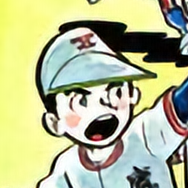}& 
        \includegraphics[width=0.105\linewidth]{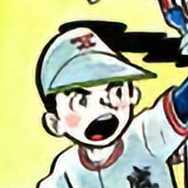}&
        \includegraphics[width=0.105\linewidth]{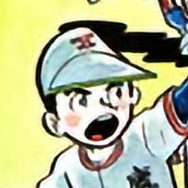}
        \\

        \includegraphics[height=0.105\linewidth]{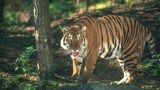}& 
        \includegraphics[width=0.105\linewidth]{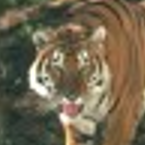}& 
        \includegraphics[width=0.105\linewidth]{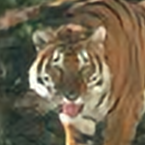}&
        \includegraphics[width=0.105\linewidth]{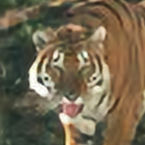}& 
        \includegraphics[width=0.105\linewidth]{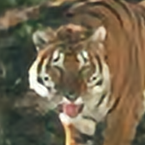}& 
        \includegraphics[width=0.105\linewidth]{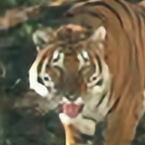}& 
        \includegraphics[width=0.105\linewidth]{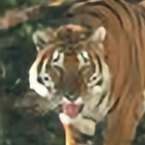}&
        \includegraphics[width=0.105\linewidth]{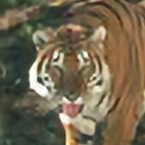}
        \\
        
        \includegraphics[height=0.105\linewidth]{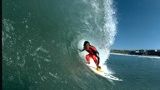}& 
        \includegraphics[width=0.105\linewidth]{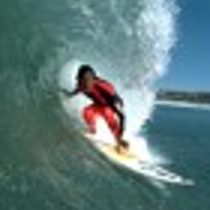}& 
        \includegraphics[width=0.105\linewidth]{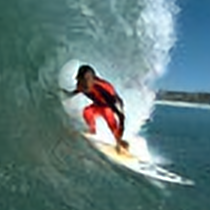}&
        \includegraphics[width=0.105\linewidth]{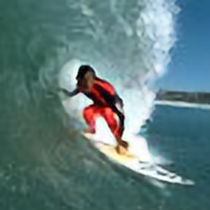}& 
        \includegraphics[width=0.105\linewidth]{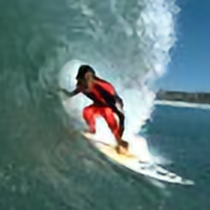}& 
        \includegraphics[width=0.105\linewidth]{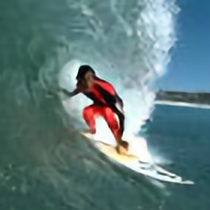}& 
        \includegraphics[width=0.105\linewidth]{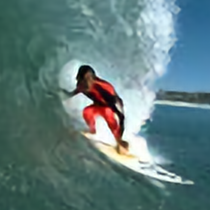}&
        \includegraphics[width=0.105\linewidth]{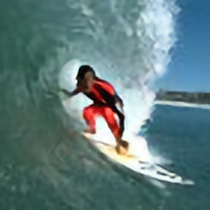}
        \\
        
        \includegraphics[height=0.105\linewidth]{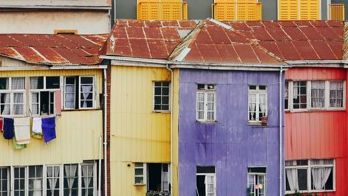}& 
        \includegraphics[width=0.105\linewidth]{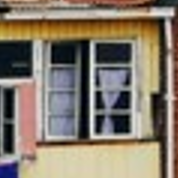}& 
        \includegraphics[width=0.105\linewidth]{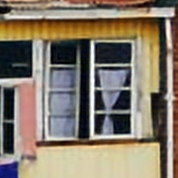}&
        \includegraphics[width=0.105\linewidth]{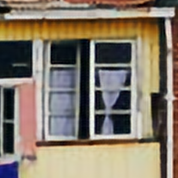}& 
        \includegraphics[width=0.105\linewidth]{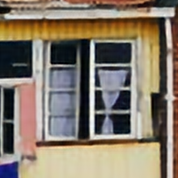}& 
        \includegraphics[width=0.105\linewidth]{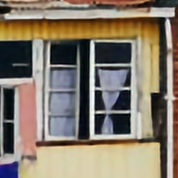}& 
        \includegraphics[width=0.105\linewidth]{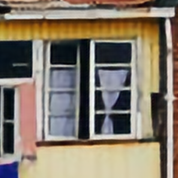}&
        \includegraphics[width=0.105\linewidth]{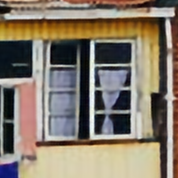}
        \\
        
        \includegraphics[height=0.105\linewidth]{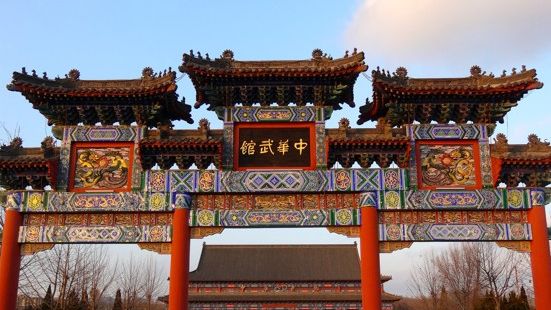}& 
        \includegraphics[width=0.105\linewidth]{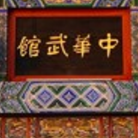}& 
        \includegraphics[width=0.105\linewidth]{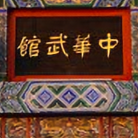}&
        \includegraphics[width=0.105\linewidth]{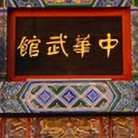}& 
        \includegraphics[width=0.105\linewidth]{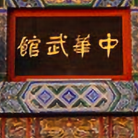}& 
        \includegraphics[width=0.105\linewidth]{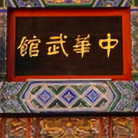}& 
        \includegraphics[width=0.105\linewidth]{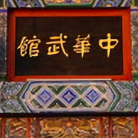}&
        \includegraphics[width=0.105\linewidth]{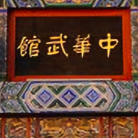}
        \\

    \end{tabular}
\caption{Qualitative comparative results of Bicubic++ and other relevant methods on the datasets~\cite{urban100,manga109,bsd100,DIV2K_1}.}
\label{fig:visual_results}
\end{figure*}

\begin{figure*}[t]
    \footnotesize
    \centering
    \addtolength{\tabcolsep}{-5pt}
    \begin{tabular}{cccccccc}
     Portion of LR & Bicubic & \textbf{Ours} & ESPCN~\cite{D2S} & XCAT~\cite{xcat} & ABPN~\cite{abpn} & XLSR~\cite{xlsr} & FSRCNN~\cite{fsrcnn}  \\
        \includegraphics[height=0.11\linewidth]{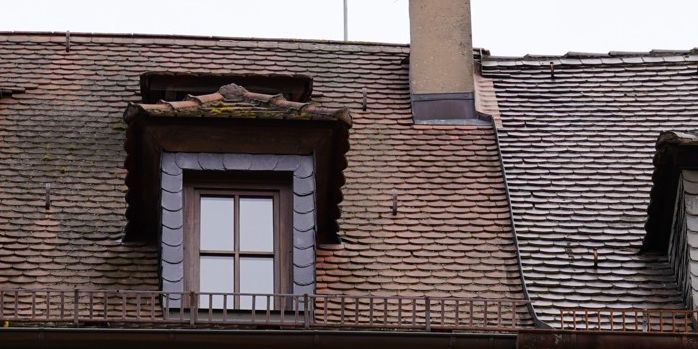}& 
        \includegraphics[width=0.11\linewidth]{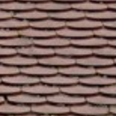}& 
        \includegraphics[width=0.11\linewidth]{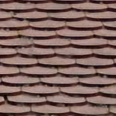}&
        \includegraphics[width=0.11\linewidth]{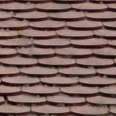}& 
        \includegraphics[width=0.11\linewidth]{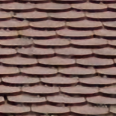}& 
        \includegraphics[width=0.11\linewidth]{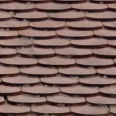}& 
        \includegraphics[width=0.11\linewidth]{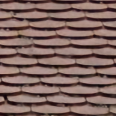}&
        \includegraphics[width=0.11\linewidth]{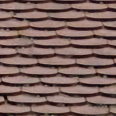}
        \\
        
        \includegraphics[height=0.11\linewidth]{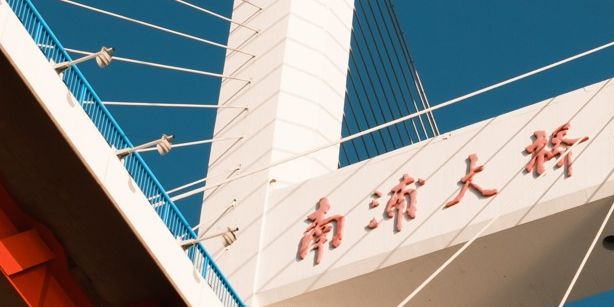}& 
        \includegraphics[width=0.11\linewidth]{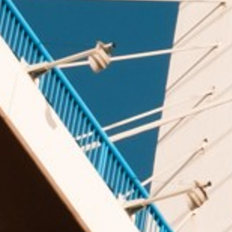}& 
        \includegraphics[width=0.11\linewidth]{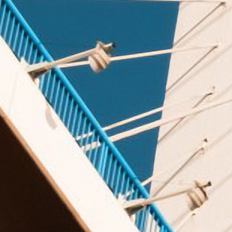}&
        \includegraphics[width=0.11\linewidth]{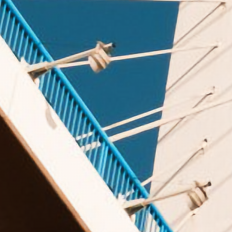}& 
        \includegraphics[width=0.11\linewidth]{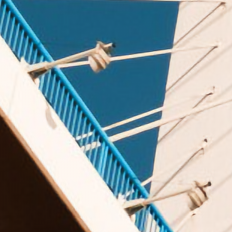}& 
        \includegraphics[width=0.11\linewidth]{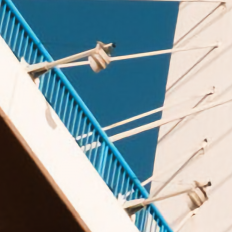}& 
        \includegraphics[width=0.11\linewidth]{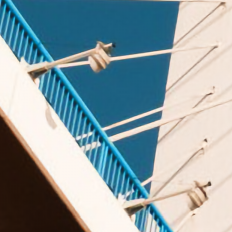}&
        \includegraphics[width=0.11\linewidth]{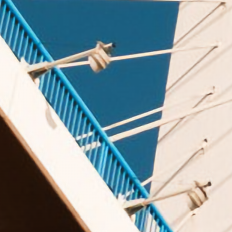}
        \\
        
        \includegraphics[height=0.11\linewidth]{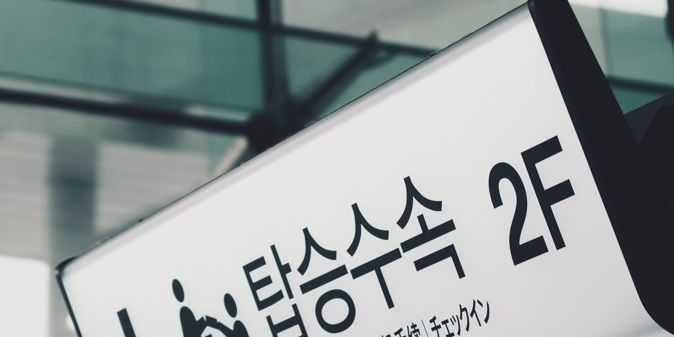}& 
        \includegraphics[width=0.11\linewidth]{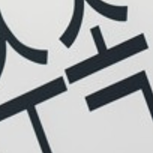}& 
        \includegraphics[width=0.11\linewidth]{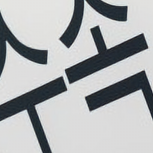}&
        \includegraphics[width=0.11\linewidth]{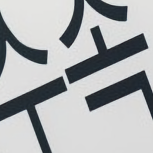}& 
        \includegraphics[width=0.11\linewidth]{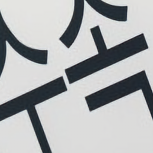}& 
        \includegraphics[width=0.11\linewidth]{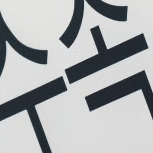}& 
        \includegraphics[width=0.11\linewidth]{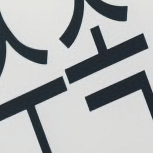}&
        \includegraphics[width=0.11\linewidth]{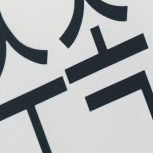}
        \\

        \includegraphics[height=0.11\linewidth]{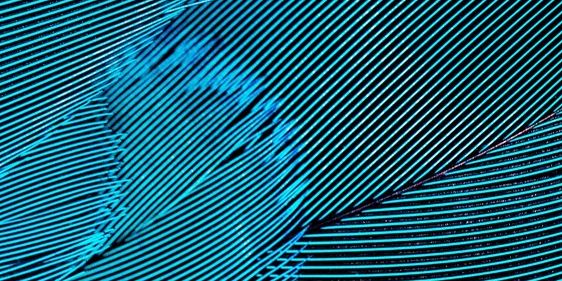}& 
        \includegraphics[width=0.11\linewidth]{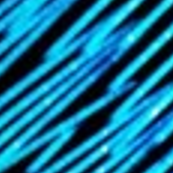}& 
        \includegraphics[width=0.11\linewidth]{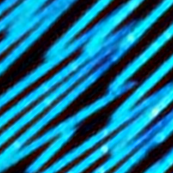}&
        \includegraphics[width=0.11\linewidth]{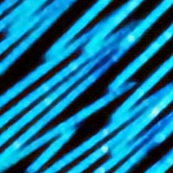}& 
        \includegraphics[width=0.11\linewidth]{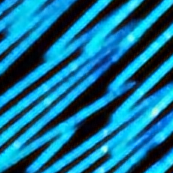}& 
        \includegraphics[width=0.11\linewidth]{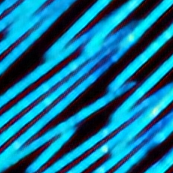}& 
        \includegraphics[width=0.11\linewidth]{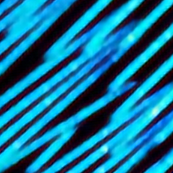}&
        \includegraphics[width=0.11\linewidth]{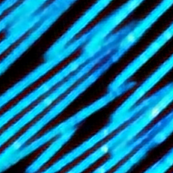}
        \\
        
        \includegraphics[height=0.11\linewidth]{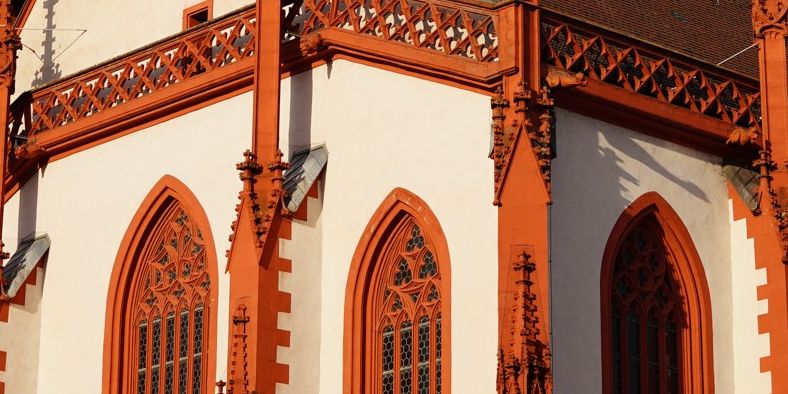}& 
        \includegraphics[width=0.11\linewidth]{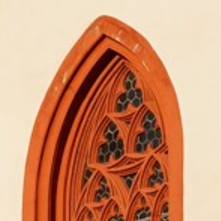}& 
        \includegraphics[width=0.11\linewidth]{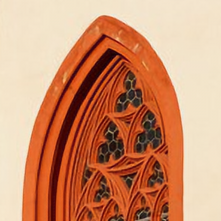}&
        \includegraphics[width=0.11\linewidth]{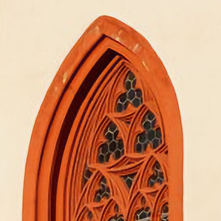}& 
        \includegraphics[width=0.11\linewidth]{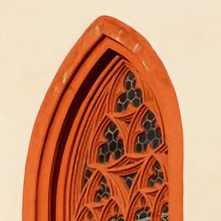}& 
        \includegraphics[width=0.11\linewidth]{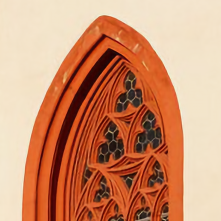}& 
        \includegraphics[width=0.11\linewidth]{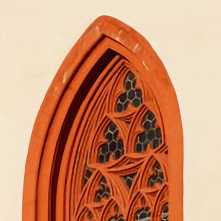}&
        \includegraphics[width=0.11\linewidth]{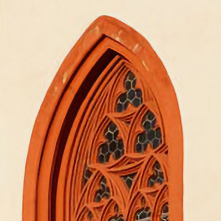}
        \\
        
        \includegraphics[height=0.11\linewidth]{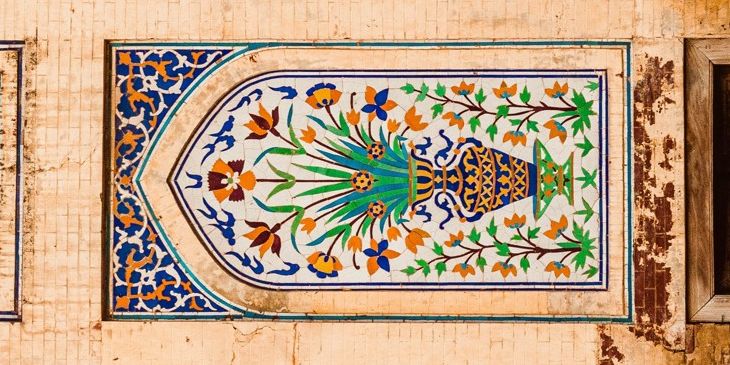}& 
        \includegraphics[width=0.11\linewidth]{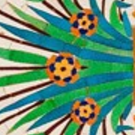}& 
        \includegraphics[width=0.11\linewidth]{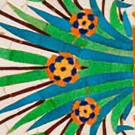}&
        \includegraphics[width=0.11\linewidth]{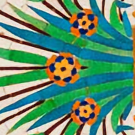}& 
        \includegraphics[width=0.11\linewidth]{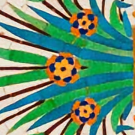}& 
        \includegraphics[width=0.11\linewidth]{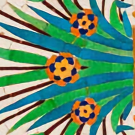}& 
        \includegraphics[width=0.11\linewidth]{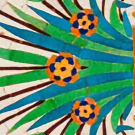}&
        \includegraphics[width=0.11\linewidth]{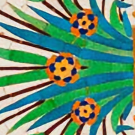}
        \\
        
       \includegraphics[height=0.11\linewidth]{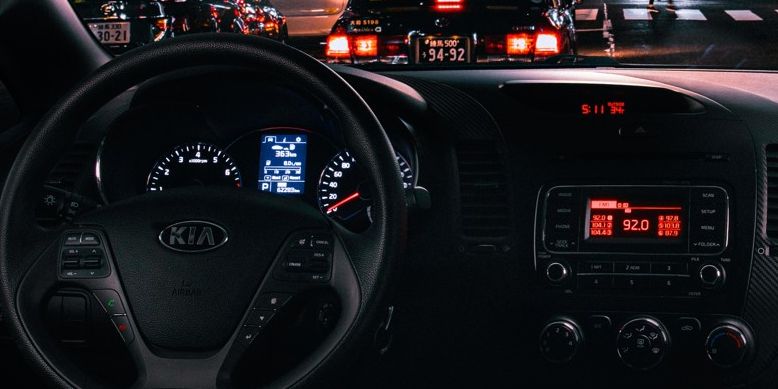}& 
        \includegraphics[width=0.11\linewidth]{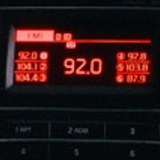}& 
        \includegraphics[width=0.11\linewidth]{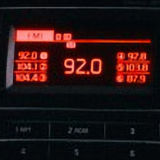}&
        \includegraphics[width=0.11\linewidth]{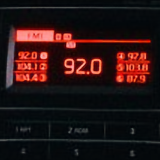}& 
        \includegraphics[width=0.11\linewidth]{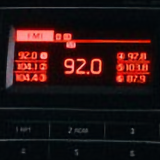}& 
        \includegraphics[width=0.11\linewidth]{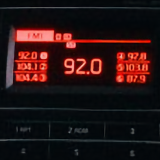}& 
        \includegraphics[width=0.11\linewidth]{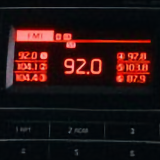}&
        \includegraphics[width=0.11\linewidth]{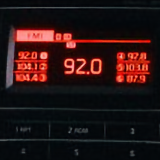}
        \\
        
        \includegraphics[height=0.11\linewidth]{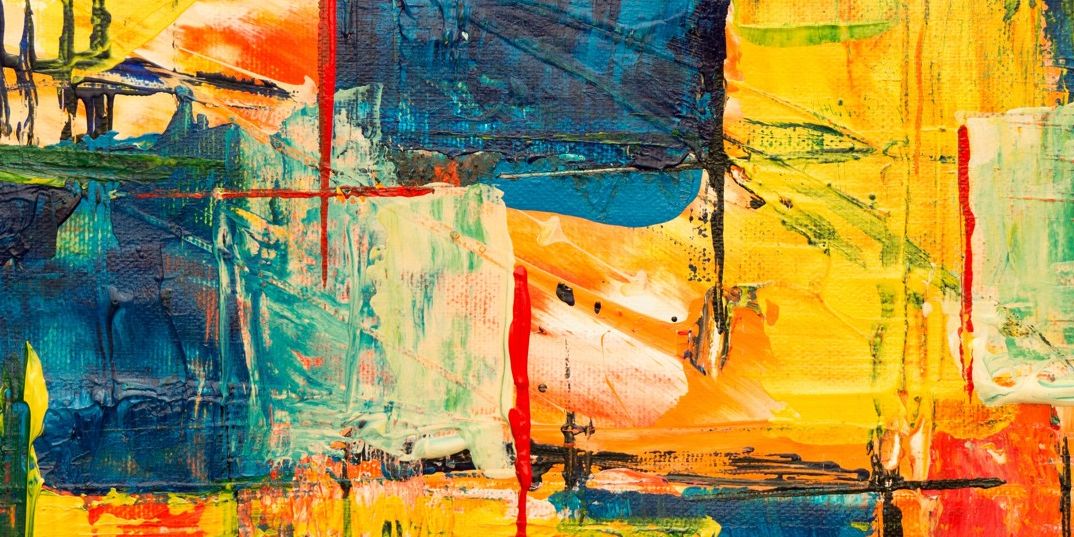}& 
        \includegraphics[width=0.11\linewidth]{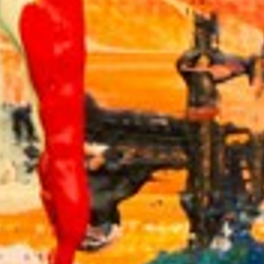}& 
        \includegraphics[width=0.11\linewidth]{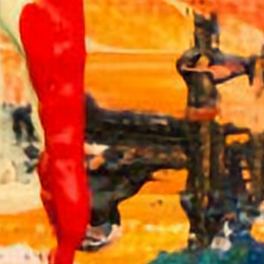}&
        \includegraphics[width=0.11\linewidth]{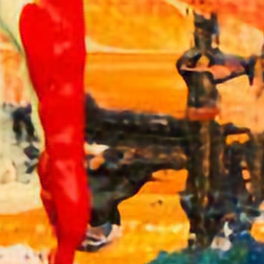}& 
        \includegraphics[width=0.11\linewidth]{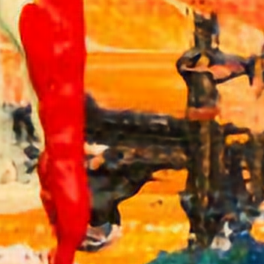}& 
        \includegraphics[width=0.11\linewidth]{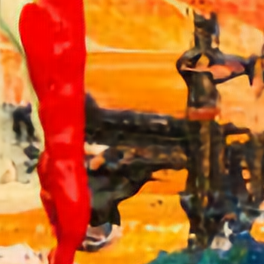}& 
        \includegraphics[width=0.11\linewidth]{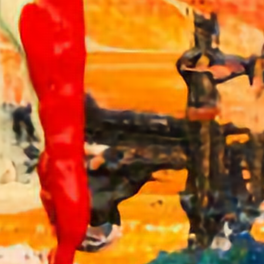}&
        \includegraphics[width=0.11\linewidth]{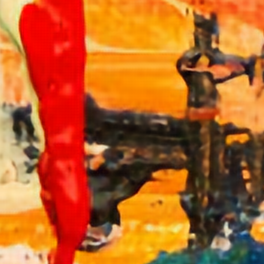}
        \\
        
        \includegraphics[height=0.11\linewidth]{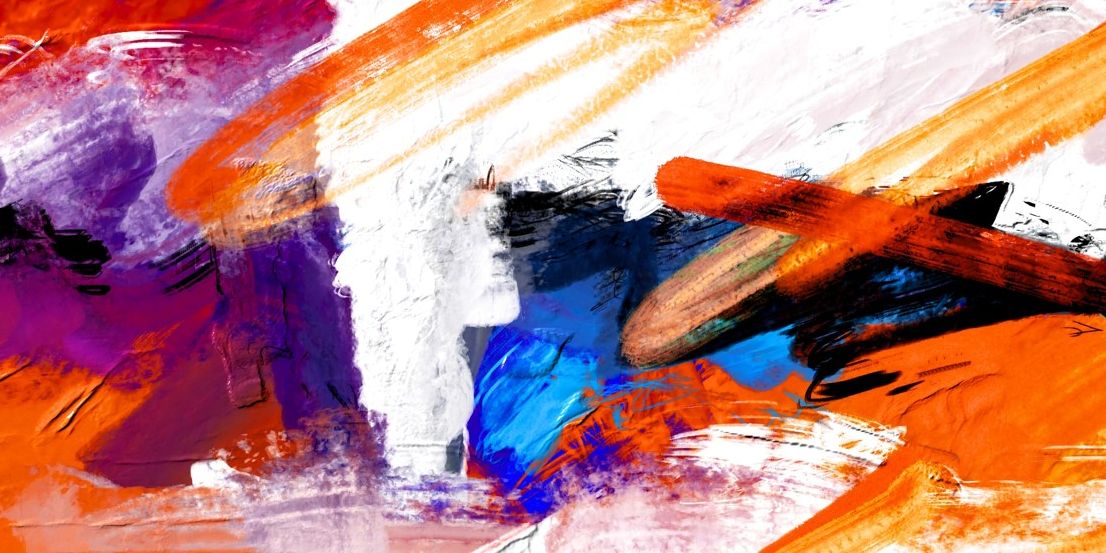}& 
        \includegraphics[width=0.11\linewidth]{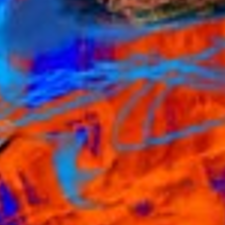}& 
        \includegraphics[width=0.11\linewidth]{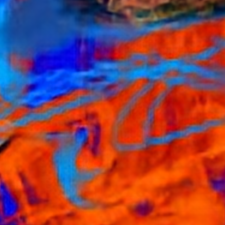}&
        \includegraphics[width=0.11\linewidth]{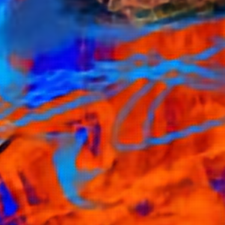}& 
        \includegraphics[width=0.11\linewidth]{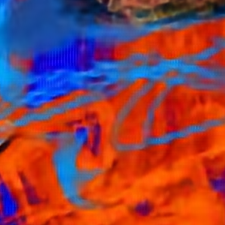}& 
        \includegraphics[width=0.11\linewidth]{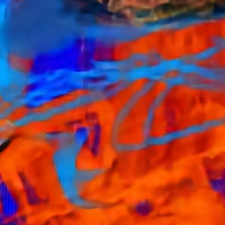}& 
        \includegraphics[width=0.11\linewidth]{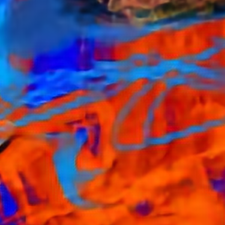}&
        \includegraphics[width=0.11\linewidth]{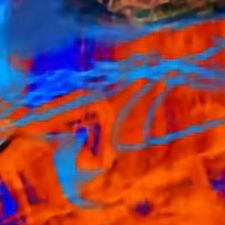}
        \\
        
        \includegraphics[height=0.11\linewidth]{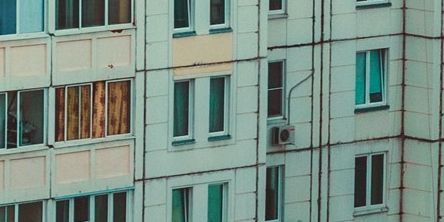}& 
        \includegraphics[width=0.11\linewidth]{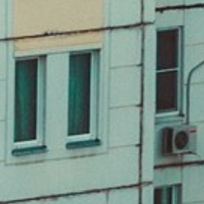}& 
        \includegraphics[width=0.11\linewidth]{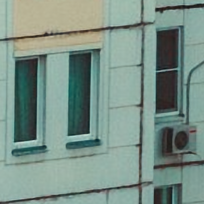}&
        \includegraphics[width=0.11\linewidth]{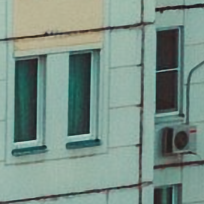}& 
        \includegraphics[width=0.11\linewidth]{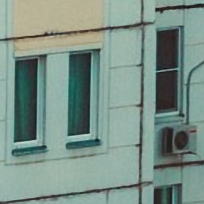}& 
        \includegraphics[width=0.11\linewidth]{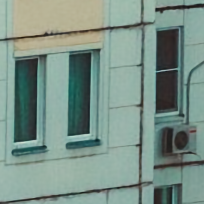}& 
        \includegraphics[width=0.11\linewidth]{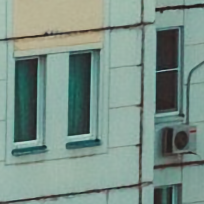}&
        \includegraphics[width=0.11\linewidth]{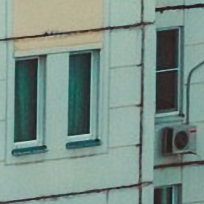}
        \\
    \end{tabular}
\caption{Qualitative comparative results of Bicubic++ and other relevant methods on the challenge test set~\cite{zamfir2023rtsr}.}
\label{fig:visual_results_2}
\end{figure*}

We provide quantitative and qualitative comparative results in~\cref{tab:top_5},~\cref{tab:psnr_comparison},~\cref{fig:visual_results}, and~\cref{fig:visual_results_2}. Bicubic++ achieves better results than the traditional Bicubic upsampling and runs much faster compared to the other relevant methods, making it a good candidate for real-time SR appplications.

Although relevant, we did not include the results of sparse coding methods like A+~\cite{A_plus} and SR-LUT~\cite{SR-LUT} since they are not compliant with the challenge requirements, either due to their slower runtime or lack of implementation in GPU.


\begin{table}[ht!]
\centering
\scriptsize
\begin{tabular}{@{}rccccc@{}}
\toprule
\multirow{2}{*}{} & \multirow{2}{*}{\begin{tabular}[c]{@{}c@{}}{Track 2} \\ {Score}\end{tabular}}  & \multirow{2}{*}{\begin{tabular}[c]{@{}c@{}}{PSNR} \\ {(RGB)}\end{tabular}} & \multirow{2}{*}{SSIM}            & \multirow{2}{*}{\begin{tabular}[c]{@{}c@{}}{PSNR} \\ {(Y)}\end{tabular}}       & \multirow{2}{*}{\begin{tabular}[c]{@{}c@{}}{Runtime} \\ {(ms)}\end{tabular}}  \\  \\ \midrule
\textbf{Aselsan Research} & \textbf{31.26} & \textbf{32.06}                 & \textbf{0.8344} & \textbf{34.56} & \textbf{1.17} \\
Team OV                   & 29.63          & 32.17                          & 0.8376          & 34.72          & 1.51          \\
ALONG                     & 28.57          & 32.18                          & 0.8367          & 34.66          & 1.66          \\
RTVSR                     & 26.89          & 32.22                          & 0.8372          & 34.77          & 1.96          \\
Noah\_TerminalVision      & 26.68          & 32.65                          & 0.8455          & 35.10          & 3.64          \\
NJUST-RTSR                & 23.51          & 32.25                          & 0.8384          & 34.90          & 2.68          \\
Antins\_cv                & 23.44          & 32.63                          & 0.8457          & 35.21          & 4.60          \\
DFCDN Team                & 22.64          & 32.07                          & 0.8371          & 34.63          & 2.25          \\
Multimedia                & 21.55          & 32.33                          & 0.8398          & 34.83          & 3.56          \\
z6                        & 20.90          & 32.59                          & 0.8446          & 35.05          & 5.47          \\
R.I.P. ShopeeVideo        & 15.67          & 32.84                          & 0.8469          & 35.30          & 13.79         \\
ECNU\_SR                  & 15.39          & 32.64                          & 0.8458          & 35.17          & 10.75         \\
Touch\_Fish               & 11.55          & 32.67                          & 0.8468          & 35.31          & 19.86         \\
P.AI.R                    & 8.66           & 32.55                          & 0.8441          & 35.04          & 30.03         \\
SEU\_CNII                 & 6.68           & 31.85                          & 0.8326          & 34.52          & 19.05         \\
diSRupt                   & 6.34           & 31.64                          & 0.8292          & 34.25          & 16.00         \\ \midrule
Bicubic              & 0.00           & 31.30                          & 0.8246          & 33.82          & 0.5           \\
Organizer's baseline~\cite{conde2023ntire_rtsr}              & 14.01           & 31.74                          & 0.8299          & 34.25          & 3.74           \\
\bottomrule
\end{tabular}
\caption{All methods scoring above Bicubic in Track 2~\cite{conde2023ntire_rtsr} and the suggested baselines. PSNR is evaluated on the competition's test set, and runtime is measured on RTX3090/3060 
with FP16 precision.} 
\label{tab:top_5}
\end{table}

\section{Conclusion and Discussion}
\label{sec:conc}
In this paper, we proposed our real-time SR network, Bicubic++. Our model learns downscaled features of the input image to increase efficiency, and is trained with our proposed three-stage pipeline where we apply global structured pruning and bias removal. We believe that our experimental results, observations, proposed architecture, and the training pipeline may help the development of real-time SR methods. We believe that Bicubic++ sets a new practical industry standard for upscaling tasks and it can be the contemporary alternative of Bicubic upscaling.

For further research, a fusion of deep learning and lookup table along with sparse representation methods can be investigated on GPUs. Furthermore, we believe that the proposed training pipeline would benefit from including reparametrization along with INT8 quantization.

\begin{table*}[]
\footnotesize
\centering
\begin{tabular}{@{}rccccccccccccc@{}}
\toprule
\multicolumn{1}{l}{} & \multicolumn{2}{c}{Set5~\cite{set5}}         & \multicolumn{2}{c}{Set14~\cite{set14}}        & \multicolumn{2}{c}{BSD100~\cite{bsd100}}      & \multicolumn{2}{c}{Urban100~\cite{urban100}}     & \multicolumn{2}{c}{Manga109~\cite{manga109}}     & \multicolumn{2}{c}{DIV2K Val~\cite{DIV2K_1}} & \multirow{2}{*}{Runtime (ms)}  \\

\multicolumn{1}{l}{} & PSNR           & SSIM            & PSNR           & SSIM            & PSNR           & SSIM            & PSNR           & SSIM            & PSNR           & SSIM            & PSNR           & SSIM            \\ \midrule
ESPCN~\cite{D2S}                & 31.56          & 0.8737          & 28.53          & 0.7833          & 27.80          & 0.7454          & 25.66          & 0.7680          & 29.29          & 0.8768          & 30.37          & 0.8351    & 9.6      \\
XCAT~\cite{xcat}                 & 31.50          & 0.8728          & 28.52          & 0.7843          & 27.81          & 0.7464          & 25.67          & 0.7697          & 29.33          & 0.8781          & 30.38          & 0.8360  & 11.7        \\
ABPN~\cite{abpn}                 & 31.93          & 0.8812          & 28.76          & 0.7910          & 27.99          & 0.7523          & 26.13          & 0.7862          & 30.14          & 0.8928          & 30.65          & 0.8421   & 15.8       \\
XLSR~\cite{xlsr}                 & 31.84          & 0.8800          & 28.74          & 0.7900          & 27.95          & 0.7509          & 25.98          & 0.7812          & 29.94          & 0.8899          & 30.59          & 0.8406   & 25.3       \\
FSRCNN~\cite{fsrcnn}               & 31.65          & 0.8762          & 28.61          & 0.7863          & 27.86          & 0.7477          & 25.81          & 0.7736          & 29.52          & 0.8814          & 30.48          & 0.8377    & 33.8      \\ \midrule
Bicubic              & 29.98          & 0.8434          & 27.30          & 0.7529          & 26.99          & 0.7180          & 24.31          & 0.7196          & 26.54          & 0.8300          & 29.33          & 0.8127        & 1.0  \\
\textbf{Bicubic++}   & \textbf{31.19} & \textbf{0.8656} & \textbf{28.35} & \textbf{0.7799} & \textbf{27.68} & \textbf{0.7431} & \textbf{25.49} & \textbf{0.7626} & \textbf{28.72} & \textbf{0.8653} & \textbf{30.24} & \textbf{0.8324} & \textbf{2.9} \\ \bottomrule
\end{tabular}
\caption{Quantitative comparative results of our method and other comparable models for $\times$3 SR. All low-resolution images in the datasets are JPEG Q90 degraded, and all PSNR values are calculated for the Y channel. Runtime is measured on RTX3070 with FP16 precision.}
\label{tab:psnr_comparison}
\end{table*}

\end{document}